\documentclass[letterpaper, 10 pt, conference]{ieeeconf}  

\IEEEoverridecommandlockouts                              

\usepackage{amsmath} 
\usepackage{amssymb}  
\usepackage{graphicx}
\usepackage{capt-of} 
\usepackage{overpic} 
\usepackage[export]{adjustbox} 
\usepackage{minibox} 
\usepackage{textcomp} 
\usepackage{tabularx}
\usepackage{multirow}
\usepackage{booktabs}
\usepackage{arydshln}
\usepackage{adjustbox} 
\usepackage{siunitx}
\usepackage{pifont}
\usepackage{stfloats}
\usepackage{makecell}
\usepackage{wrapfig}
\usepackage{url}
\usepackage{xspace}
\usepackage{csquotes} 
\usepackage{silence}
\usepackage{subcaption}
\usepackage{nicematrix}  
\usepackage[normalem]{ulem}
\usepackage{tikz}

\usepackage{algpseudocode,algorithm,algorithmicx}

\algrenewcommand\algorithmicrequire{\textbf{Precondition:}}
\algrenewcommand\algorithmicensure{\textbf{Postcondition:}}

\usepackage{cite}

\makeatletter
\DeclareRobustCommand\onedot{\futurelet\@let@token\@onedot}
\def\@onedot{\ifx\@let@token.\else.\null\fi\xspace}
\def\eg{\emph{e.g}\onedot}

\def\ie{\emph{i.e}\onedot}

\def\etal{\emph{et al}\onedot}
\makeatother

\DeclareSIUnit{\million}{\text{Mio.}}

\makeatletter
\let\NAT@parse\undefined
\makeatother

\usepackage[pagebackref,colorlinks=false,linkcolor=blue,citecolor=blue]{hyperref}
\usepackage{cleveref}

\title{\LARGE \bf
MapDiffusion: Generative Diffusion for Vectorized Online
HD Map Construction and Uncertainty Estimation in Autonomous Driving
}

\author{Thomas Monninger$^{1,2}$, Zihan Zhang$^{3,*}$, Zhipeng Mo$^{1}$, Md Zafar Anwar$^{1}$, Steffen Staab$^{2,4}$, Sihao Ding$^{1}$
\thanks{$^{1}$Mercedes-Benz Research \& Development North America, Sunnyvale, CA, USA (email: thomas.monninger@mercedes-benz.com)}%
\thanks{$^{2}$University of Stuttgart, Institute for Artificial Intelligence, Stuttgart, Germany (email: steffen.staab@ki.uni-stuttgart.de)}%
\thanks{$^{3}$University of California, San Diego, La Jolla, CA, USA}%
\thanks{$^{4}$University of Southampton, Southampton, United Kingdom}%
\thanks{$^{*}$Work was done during an internship at Mercedes-Benz Research \& Development North America.}%
}

\newcommand\copyrighttext{\footnotesize \textcopyright~2025 IEEE. Personal use of this material is permitted.  Permission from IEEE must be obtained for all other uses, in any current or future media, including reprinting/republishing this material for advertising or promotional purposes, creating new collective works, for resale or redistribution to servers or lists, or reuse of any copyrighted component of this work in other works.
}%

\newcommand\copyrightnotice{%
    \begin{tikzpicture}[remember picture,overlay]%
     \node[anchor=south, xshift=0pt, yshift=12pt] at (current page.south)%
     {\fbox{\parbox{\dimexpr\textwidth-\fboxsep-\fboxrule\relax}{\copyrighttext}}};%
     \end{tikzpicture}%
}

\begin{document}
\maketitle
\pagenumbering{arabic}

\begin{abstract}
Autonomous driving requires an understanding of the static environment from sensor data. 
Learned Bird's-Eye View (BEV) encoders are commonly used to fuse multiple inputs,
and a vector decoder predicts a vectorized map representation from the latent BEV grid.
However, traditional map construction models provide deterministic point estimates, failing to capture uncertainty and the inherent ambiguities of real-world environments, such as occlusions and missing lane markings.
We propose MapDiffusion, a novel generative approach that leverages the diffusion paradigm to learn the full distribution of possible vectorized maps.
Instead of predicting a single deterministic output from learned queries, MapDiffusion iteratively refines randomly initialized queries, conditioned on a BEV latent grid, to generate multiple plausible map samples. 
This allows aggregating samples to improve prediction accuracy and deriving uncertainty estimates that directly correlate with scene ambiguity.
Extensive experiments on the nuScenes dataset demonstrate that MapDiffusion achieves state-of-the-art performance in online map construction, surpassing the baseline by 5\% in single-sample performance.
We further show that aggregating multiple samples consistently improves performance along the ROC curve, validating the benefit of distribution modeling.
Additionally, our uncertainty estimates are significantly higher in occluded areas, reinforcing their value in identifying regions with ambiguous sensor input.
By modeling the full map distribution, MapDiffusion enhances the robustness and reliability of online vectorized HD map construction, enabling uncertainty-aware decision-making for autonomous vehicles in complex environments.
\end{abstract}


\section{Introduction}
The safe operation of an autonomous driving system requires an accurate and complete representation of the static infrastructure surrounding the vehicle.{\copyrightnotice}
This map representation must be derived in real-time from sensor information (\ie, online map construction) to react to the current real-world scenario.
Also, most autonomous driving systems require the map in vectorized form for use in their planning system, since a vectorized representation provides instance-level information and spatial consistency \cite{vectormapnet_2023, maptrv2_2023, gao2020vectornet}
However, the task of vectorized online map construction is inherently challenging due to the ambiguity of the real world.
A wide lane may be a single lane or two lanes with missing markings, intersections often lack explicit lane definitions, and construction zones introduce temporary changes that may contradict the original lane layout.
Additionally, occlusions caused by other vehicles or roadside obstacles increase ambiguity.
Committing to a single interpretation in such scenarios can be misleading and potentially unsafe.

Previous approaches to online mapping perform deterministic construction of a map from the provided sensor data.
For this, they commonly use learned Bird's-Eye View (BEV) encoders to fuse information from multiple camera views into a joint latent space.
For the decoder, initial approaches focused on predicting a raster map representation \cite{liftsplatshoot_2020, bevformer_2022, monninger2024tempbev, petrv2_2023}, while more recent approaches directly predict vectorized representations \cite{vectormapnet_2023, maptrv2_2023, streammapnet_2024}.
In both cases, previous deterministic models commit to a single interpretation for a given input.
In an ambiguous traffic scene, this map prediction may be incorrect, leading to unsafe decisions.
We argue that capturing the full distribution is required to consider all plausible map configurations in downstream decision-making.
While a few works investigate diffusion for map construction, unlike our work, they either operate on raster representations \cite{jia2024diffmap, le2024diffusion} or just do refinement of initial proposals \cite{Chen2023PolyDiffuse}.
Instead, we propose the use of full generative diffusion for vectorized High-Definition (HD) map construction.
We introduce a novel graph diffusion decoder that denoises randomly initialized queries conditioned on camera features from the latent BEV grid.
MapDiffusion learns the full vector map distribution and therefore can generate plausible samples to capture the ambiguity of the real world.
Furthermore, we show that the variance between the sampling results can serve as a measure of uncertainty in the constructed online map, providing an effective way to capture ambiguity in the real world.

\begin{figure}[t!]
    \vspace{3pt}  
    \includegraphics[width=1.0\columnwidth]{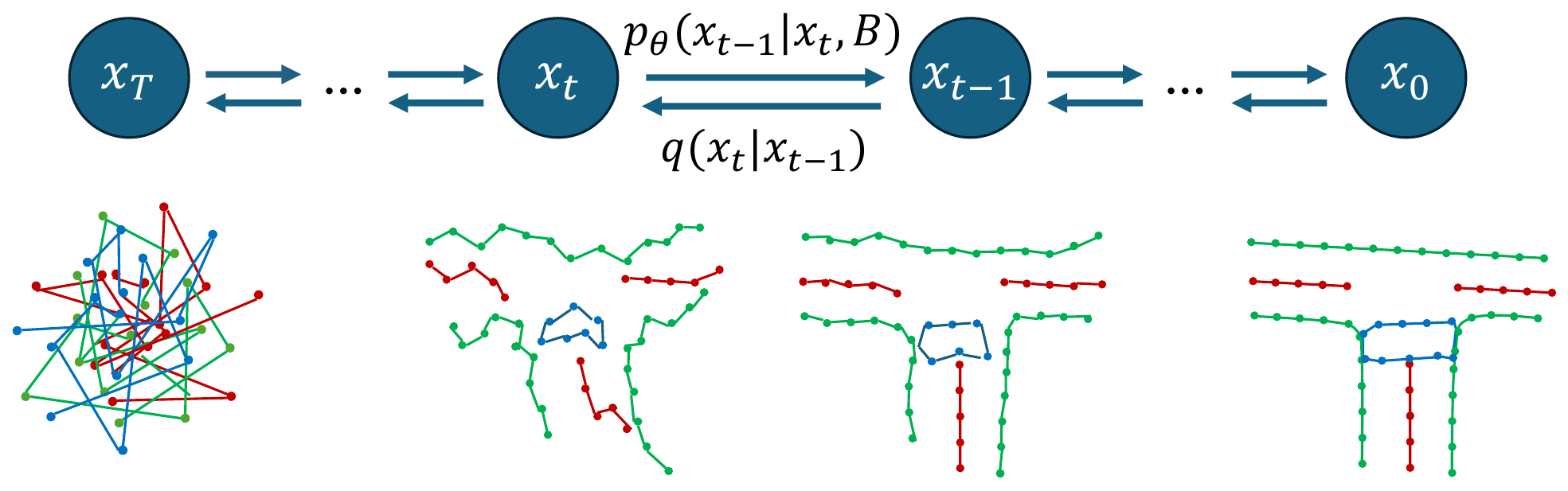}
    \caption{Overview of the diffusion forward and backward processes on vectorized HD maps used in MapDiffusion.}
    \label{fig:vectorized_diffusion}
\end{figure}

The primary contributions of this paper are as follows:
\begin{itemize}
\item We propose the use of generative diffusion for the task of online vectorized HD map construction and implement a new model, MapDiffusion, that performs denoising conditioned on a latent BEV grid.
\item We investigate the variance of the sampling output by interpreting it as a measure of uncertainty and find a significant increase of \SI{31}{\percent} in occluded areas.
\item We conduct extensive experiments on the nuScenes dataset, demonstrating a 5~\% relative improvement in single-sample performance and further improvements by aggregating multiple samples.
\end{itemize}

\section{Related Work}\label{sec:related_work}

\subsection{Online HD Map Construction}
In autonomous driving, static elements such as roads, lane dividers, and pedestrian crossings are typically represented in a map.
Recent work has made notable progress in constructing a map representation online directly from sensor data.
Typically, a learned BEV encoder is used to fuse information from multiple camera views into a joint representation.
The first works on learned BEV encoders predict a raster map by treating it as a segmentation task, \ie, a pixel-wise classification of map elements \cite{liftsplatshoot_2020, bevformer_2022, monninger2024tempbev, petrv2_2023}.

Starting with VectorMapNet \cite{vectormapnet_2023}, more recent approaches construct end-to-end HD maps by directly predicting the vectorized map elements.
MapTR \cite{maptr_2023} addresses the ambiguity of selecting a discrete set of points to model geometries.
It employs permutation-equivalent modeling to stabilize the learning process. 
StreamMapNet \cite{streammapnet_2024} uses a powerful 6-layer transformer decoder that performs temporal aggregation by streaming queries from the previous frame.
AugMapNet \cite{monninger2025augmapnet} adds dense spatial supervision for improved structure of the latent space.
SQD-MapNet \cite{wang2024stream} injects a few noise-perturbed GT queries from the previous frame during training.
A query denoising module is added to improve temporal consistency.
We use the general idea of query denoising in the context of a diffusion framework.
In contrast, during our training, we only use noise-perturbed GT queries from the current frame and execute the full decoder iteratively to learn sampling from random Gaussian noise.

\subsection{Diffusion Models}
Generative modeling can generate complex objects in a domain, \eg, high-fidelity image synthesis, video generation, and natural language processing \cite{ho2020denoising,song2019generative,dhariwal2021diffusion,chen2023diffusiondet}.
The diffusion process can be controlled by conditioning on additional information \cite{song2021scorebased, ramesh2022hierarchical}. 
We leverage generative modeling to sample from a probability distribution over vectorized maps conditioned on sensor data encoded in a BEV grid $B$.

Denoising Diffusion Probabilistic Models (DDPMs) \cite{ho2020denoising} model a Markovian forward process, $q$, that transforms $x$ into Gaussian noise over multiple diffusion time steps $t$.
\Cref{fig:vectorized_diffusion} visualizes this for a vectorized HD map with $x_0$ being the vectorized map and $x_T$ being polylines with random coordinates.
A denoising prediction network with weights $\theta$ learns the reverse process \( p_\theta(x_t, t, B) \), where the latent BEV grid $B$ serves as condition to guide the prediction.
Once trained on a distribution, the diffusion model can generate samples from that distribution.

\subsection{Diffusion Models for Mapping}
As a first application, the diffusion paradigm has been used to generate 3D occupancy maps \cite{zheng2024occworld, wang2024occgen, li2024uniscene, reed2024online}, capturing the three-dimensional geometric structure of the surroundings.
Other works \cite{gu2024generating, ruiz2024lane,wang2024radiodiff} use diffusion models conditioned on aerial images or other geospatial context to generate semantic map layers for various use cases.
More recently, works have begun to explore diffusion models to construct online raster maps from on-road camera views.
DiffMap \cite{jia2024diffmap} leverages a latent diffusion model and enhances the generated raster map by integrating structured priors inherent in map segmentation masks.
DifFUSER \cite{le2024diffusion} extends the diffusion paradigm to both 3D object detection and raster map prediction.
In contrast to the above works, our approach applies the diffusion paradigm to directly predict vectorized map elements.

A new research topic is diffusion-based generation of vector representations.
DiffusionDet \cite{chen2023diffusiondet} and DiffBEV \cite{zou2024diffbev} apply the diffusion paradigm to object detection and generate vectorized bounding boxes.
They condition the diffusion process on the image and BEV space, respectively.
HouseDiffusion \cite{shabani2023housediffusion} performs diffusion to generate vectorized floor plans, using similar techniques to our work, but for a different learning task.
To our knowledge, the only work that uses diffusion for online vectorized map construction is PolyDiffuse \cite{Chen2023PolyDiffuse}.
Its Guided Set Diffusion Model uses a guidance network to manage noise injection and maintain unique representations for the diffusion model, enabling accurate polygonal shape reconstruction of floor plans and HD maps. 
However, they use diffusion as a refinement step on top of coarse predictions from existing map construction models, correcting structural errors and enhancing the accuracy of the predicted polylines. 
Unlike MapDiffusion, their overall accuracy remains heavily dependent on the performance of the baseline model, and their method cannot be used to generate diverse samples from the learned distribution.

\subsection{Uncertainty Estimation}
Gu \etal \cite{gu2024producing} extend methods for online map construction with uncertainty estimation.
Instead of predicting the vectorized coordinates, they predict the parameters of a Laplace distribution for each polyline point.
To show the benefit of predicting a map distribution, they evaluate its use in a trajectory prediction model and find up to \SI{15}{\percent} improved prediction performance.
We follow this argument and produce uncertainty estimates from the sampling variance of our diffusion process.
By considering multiple samples, we not only rely on one set of predicted polylines, yielding denser spatial uncertainty estimates.

Diffusion models have successfully been used for uncertainty estimation in other domains.
CARD \cite{han2022card} is a diffusion-based approach that uses conditional generative models to uncover predictive distributions, hence capturing the uncertainty.
Du and Li \cite{du2023diffusion} also use diffusion for uncertainty estimation and apply this to active domain adaptation.

In the context of trajectory prediction, uncertainty is inherently present in the task due to its multi-modal distribution.
MotionDiffuser \cite{jiang2023motiondiffuser} uses controllable diffusion to sample plausible trajectories.
To the best of our knowledge, we are the first to leverage diffusion-based sampling to estimate the uncertainty of online map construction.

\section{Approach}

\subsection{Problem Statement} \label{sec:problem}
Let $I=\{i_1, \ldots, i_m\}$ be the set of image frames from the $m$ monocular cameras mounted on the ego vehicle.
Moreover, for a given scene, let $\mathcal{P}_{\mathrm{div}}$, $\mathcal{P}_{\mathrm{bound}}$, and $\mathcal{P}_{\mathrm{ped}}$ be the sets of polylines representing lane dividers, lane boundaries, and pedestrian crossings, respectively, with a polyline, $P = \left[ (x_{j},y_{j}) \right]_{j=1}^{N_{P}}$, being a sequence of $N_{P}$ points.
Let $\mathcal{M} = \left( \mathcal{P}_{\mathrm{div}}, \mathcal{P}_{\mathrm{bound}}, \mathcal{P}_{\mathrm{ped}} \right)$ be the local HD map with the ego vehicle at the origin.
The goal is to find a function $f$ that returns an estimate of the local HD map, $\mathcal{\hat{M}}$, for a given set of image frames ${I}$, \ie $\mathcal{\hat{M}} = f \left( I \right)$.
Additionally, the goal is to provide a function $\mathcal{U}$ that provides an uncertainty estimate for $\mathcal{\hat{M}}$ at a Cartesian location $(x,y) \in \mathbb{R}^2$.
This uncertainty estimate provides a fuzzy, qualitative indicator of confidence for the predicted map at each location in the scene based on the perceived ambiguity at that location.

\subsection{MapDiffusion} \label{sec:approach}
\begin{figure*}
    \centering
    \vspace{0.4em}  
    \includegraphics[width=1.0\textwidth]{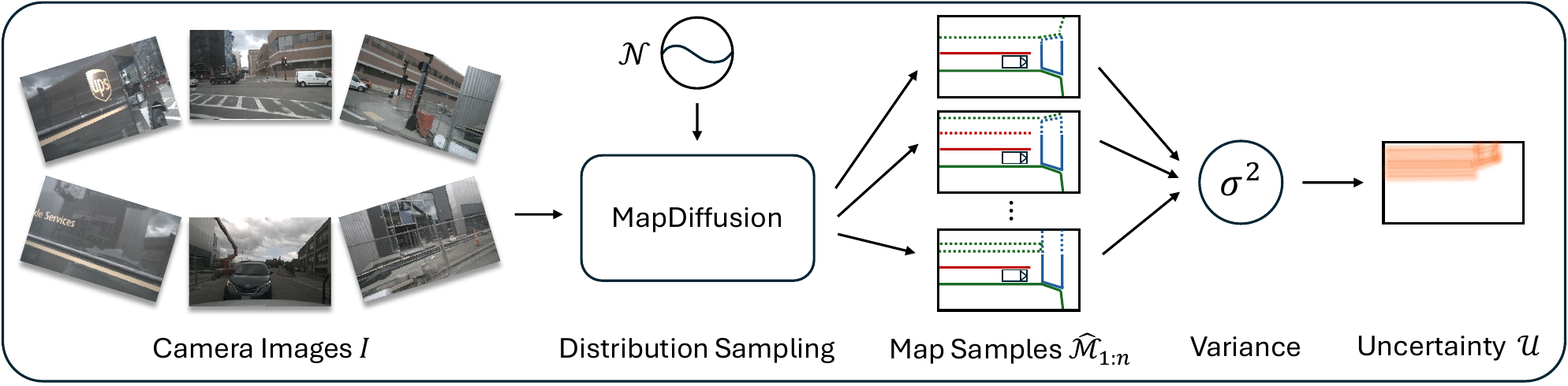}
    \caption{Illustration of MapDiffusion approach based on schematic traffic scene with an occluded camera view on the ego's left. MapDiffusion uses generative diffusion to predict samples of map distribution from camera images. The samples show plausible predictions for the occluded area on the ego's left (differences are indicated with dashed lines). The variance across samples is used as spatial uncertainty estimation, yielding high uncertainty in the occluded area (in orange on the right).}
    \label{fig:approach}
\end{figure*}
We propose MapDiffusion, a novel model that leverages generative diffusion to sample map predictions.
An overview of our approach is shown in \Cref{fig:approach}.
MapDiffusion can generate samples from the learned map distribution $\mathcal{M}$.
The variance across generated map samples $\hat{\mathcal{M}}$ serves as spatial uncertainty estimate $\mathcal{U}$.
In the figure, the camera view to the left is blocked by a delivery truck.
MapDiffusion can sample plausible map configurations with high variance in the occluded area.
Consequently, $\mathcal{U}$ suggests a high uncertainty in that area.
The following sections cover various aspects of our MapDiffusion approach.

\subsubsection{Model Architecture} \label{sec:model}

MapDiffusion uses StreamMapNet \cite{streammapnet_2024} as reference architecture; the high-level architecture of both models is shown in \Cref{fig:comparison_smn_mapdiffuse}.
Both use a learned BEV encoder to generate a latent representation of the camera features.
We leverage a DETR-style transformer decoder \cite{carion2020end} that performs query refinement conditioned on the latent representation of the BEV grid.
We adapt the decoder to a diffusion framework such that the denoising decoder starts with random polylines as input and progressively performs denoising through an iterative refinement process. 

\subsubsection{Training Process and Noise Scheduler}
\begin{figure*}
    \centering
    \vspace{0.4em}  
    \begin{subfigure}[t]{0.315\textwidth}
        \centering
        \includegraphics[trim=0cm 0cm 0cm 0cm, clip, width=\textwidth]{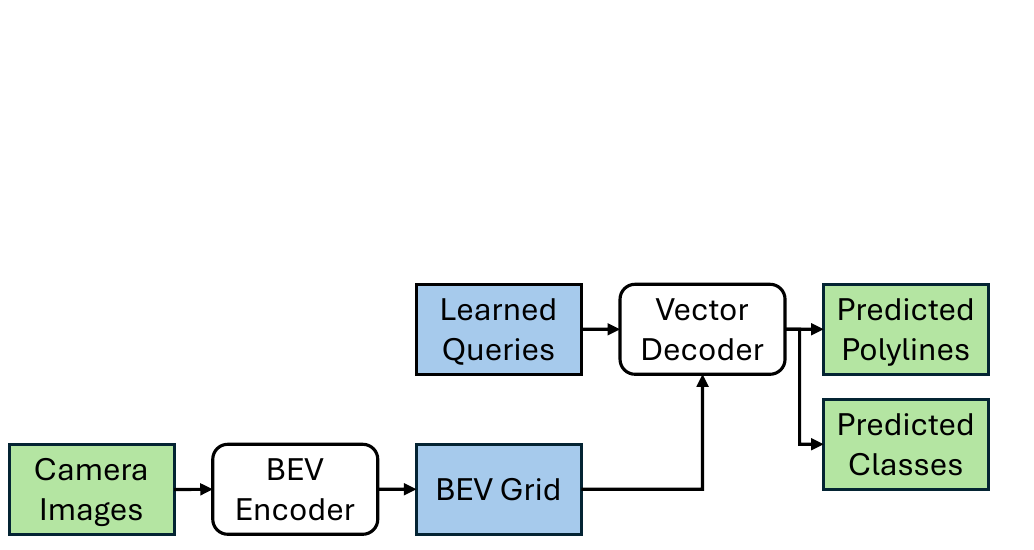}
        \caption{StreamMapNet}
        \label{fig:comparison_smn}
    \end{subfigure}
    \hfill
    \begin{subfigure}[t]{0.315\textwidth}
        \centering
        \includegraphics[trim=0cm 0cm 0cm 0cm, clip, width=\textwidth]{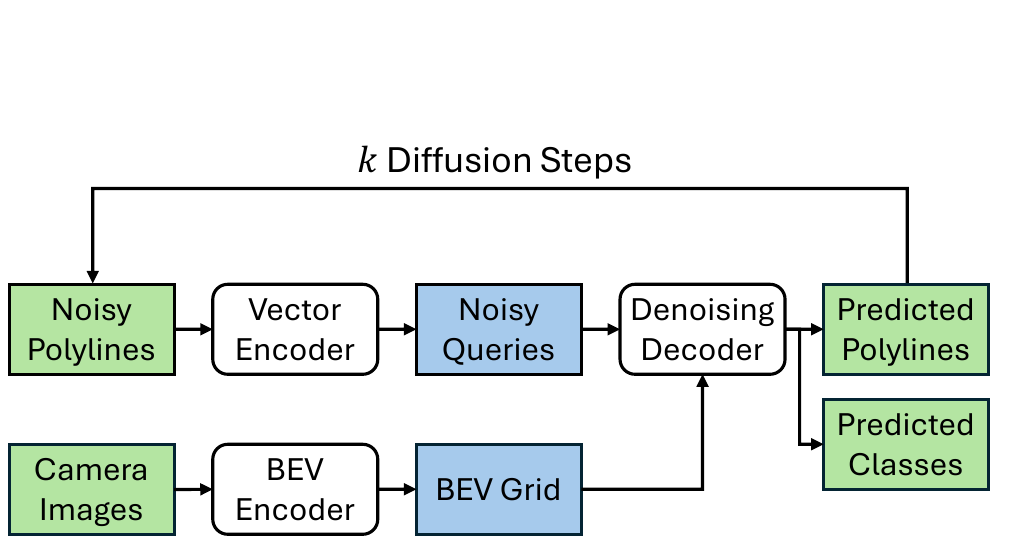}
        \caption{MapDiffusion (Inference)}
        \label{fig:comparison_mapdiffusion_inference}
    \end{subfigure}
    \hfill
    \begin{subfigure}[t]{0.315\textwidth}
        \centering
        \includegraphics[trim=0cm 0cm 0cm 0cm, clip, width=\textwidth]{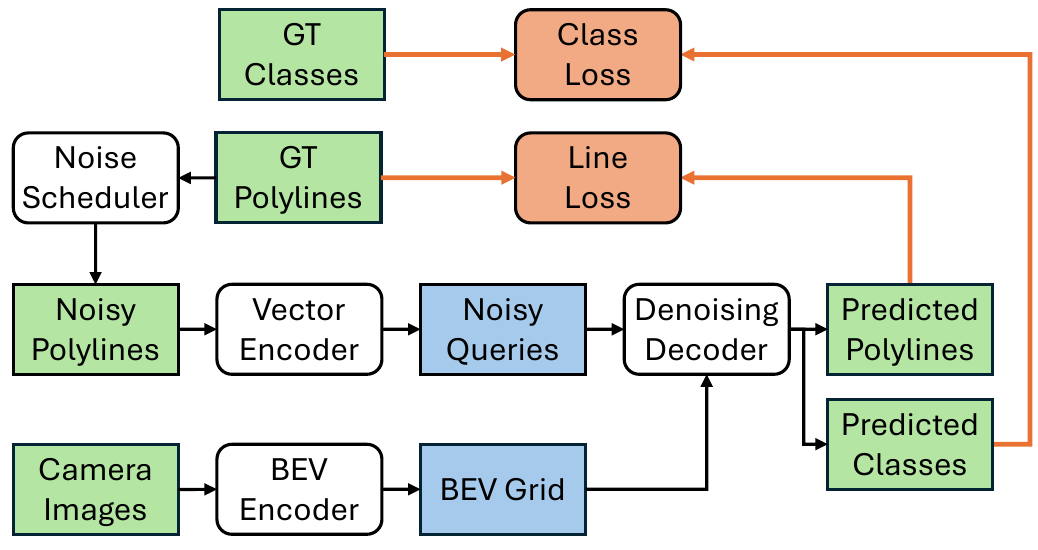}
        \caption{MapDiffusion (Training)}
        \label{fig:comparison_mapdiffusion_train}
    \end{subfigure}
   \caption{Overview of StreamMapNet, MapDiffusion during inference, and MapDiffusion during training, with box colors indicating explicit representations (green), latent representations (blue), computation modules (white), and losses (orange).}
    \label{fig:comparison_smn_mapdiffuse}
\end{figure*}

\Cref{fig:comparison_mapdiffusion_train} shows an overview of the training process.
During training, a noise scheduler performs the forward process $q$.
The Noise Scheduler determines the noise added to the GT at each diffusion time step $t$.
As visualized in \Cref{fig:vectorized_diffusion}, we perform $q$ and $p_\theta$ in vector space.
The denoising decoder uses an embedding of the time step $t$ as an additional input to condition its prediction on the noise step. 
It is trained to minimize the error between $p_\theta ( q ( x_0 , t), t, B)$ and $x_0$ for all $t \in [0, T]$, which is visualized \enquote{Line Loss} in \Cref{fig:comparison_mapdiffusion_train}. 
Prediction of the class score is excluded from the diffusion process and instead performed in the final step, which enables diffusion to happen purely in the vector space.
The overall training is a joint optimization of the diffusion process and the classification task. 

\subsubsection{Diffusion Conditioning}
MapDiffusion conditions the diffusion process on a latent BEV grid $B$ to guide the denoising process by camera features. 
This ensures that the generated map prediction is not only a plausible map, but also consistent with what is visible in the camera images $I$.
To integrate this additional context, we employ deformable cross-attention \cite{zhu2021deformable} to have the denoising decoder query the latent BEV grid.

\subsubsection{Query Padding}
The DETR-style decoder \cite{carion2020end} operates with a fixed number of $l$ queries, each corresponding to either a map element in the output, or the \enquote{no object} class.
During the training of MapDiffusion, the GT map elements serve as a starting point, and the forward process $q$ is applied to initialize the queries. 
However, the GT contains a variable number of map elements, typically fewer than $l$, so padding is required for the queries.
Padding with Gaussian noise performs best based on our ablation study (see \Cref{sec:padding_ablation}).

\subsubsection{Temporal Aggregation in Decoder}
MapDiffusion uses the BEV-space temporal aggregation on the encoder side from StreamMapNet \cite{streammapnet_2024}.
StreamMapNet additionally uses temporal aggregation in the vector decoder by incorporating refined queries from the previous temporal step into the second layer of the decoder.
In our diffusion decoder, we opt not to use the refined queries from the previous temporal step so that map predictions are generated purely from random noise. 
While this misses out on past object-space information to improve the consistency of the predicted map elements over time, it simplifies the architecture and preserves the original concept of diffusion.

\subsubsection{Sampling}
The sampling process follows Denoising Diffusion Implicit Model (DDIM) \cite{song2020denoising}, which allows generating high-quality map predictions with only a few diffusion steps. 
The sampling process is shown in \Cref{fig:comparison_mapdiffusion_inference}.
The BEV encoder at the bottom is only calculated once for efficiency.
The resultant latent BEV grid is used to condition the denoising decoder.
The denoising decoder refines the vectorized polyline prediction through $k$ diffusion steps.
Additionally, filtering of the output is performed during inference. Specifically, outputs with a score below $\tau$ are dropped and randomly initialized for the next inference step. This prevents the model from having to deal with suboptimal initializations.

\subsection{Aggregating Samples for Refined Prediction and Uncertainty Estimation} \label{sec:aggregation}
During inference, for a given set of camera images $I$, we sample $n$ map predictions from the map distribution.
We use different aggregation strategies to get to a refined prediction and to generate an uncertainty estimation.
To get the distribution of the predicted local HD map, we generate $n$ samples of map predictions, denoted as $\hat{\mathcal{M}}_i$, for $i \in [1, 2, \dots, n]$.
It is important to note that the number of samples $n$ is different from the number of diffusion steps $k$ performed to generate one sample.
We present an aggregation strategy to get a refined prediction as well as an uncertainty estimate. 

\subsubsection{Sample Aggregation as Refined Prediction} \label{sec:refined}
We aggregate $n$ samples to obtain a refined prediction from a map distribution. 
Since the aggregation of sets of polylines is non-trivial, we perform aggregation in raster space, which is sufficient to demonstrate our point.
For each predicted sample $i$ from $n$ samples and class $c \in C$, let $\mathcal{\hat{P}}_c^i$ denote the set of predicted polylines corresponding to class $c$ and let $\mathcal{S}_c^i$ represent the associated confidence scores. 
A polyline indexed by $j$, $\mathcal{\hat{P}}_c^i[j]$, is converted into a rasterized map, and each location is weighted by its corresponding score $\mathcal{S}_c^i[j]$. All polylines for class $c$ are summed to produce the weighted raster $\mathcal{R}_c^i \in \mathbb{R} ^{H \times W}$:
\begin{equation}
\mathcal{R}_c^i(x, y) = \sum_{j = 1}^{|\mathcal{\hat{P}}_c^i|} \mathcal{S}_c^i[j]\, \text{Rasterize}(\mathcal{\hat{P}}_c^i[j]).
\end{equation}

Aggregation of the class probability distributions requires spatial smoothness, so a Gaussian kernel, $G \in \mathbb{R}^{g \times g}$, is applied to the weighted raster $\mathcal{R}_c^i$.
Due to potential overlap of different polylines, the resulting values are clipped to the range $[0,1]$ to obtain a class probability map $\mathcal{D}_c^i$:
\begin{equation}
\mathcal{D}_c^i(x, y) = \min(1, G \ast \mathcal{R}_c^i(x, y)),
\end{equation}
where $\ast$ denotes the convolution operation.

The aggregated class probability distribution is given by the mean probability per class at location $(x, y)$ as:
\begin{equation}
\mathcal{D}_c(x, y) = \frac{1}{n} \sum_{i=1}^{n} \mathcal{D}_c^i(x, y).
\end{equation}

To generate a refined prediction from the distribution, the scores are thresholded with a binarization threshold $b$.

\subsubsection{Sample Variance as Uncertainty Estimation} \label{sec:uncertainty}
According to the problem statement, we need an uncertainty estimation that captures the ambiguity of the real world.
Our diffusion model can generate diverse samples that capture the multi-modality of the map distribution.
Given $n$ samples from our model for a given set of images $I$, we can leverage the sample variance to compute a spatial uncertainty.
For simplicity, we construct an uncertainty estimate at each spatial location $(x, y)$ of a predefined grid, $\mathcal{U} \in \mathbb{R} ^{H \times W}$, based on the total variance across class scores at that location.
Specifically, we first compute the per-location variance, $\sigma^2_c$, by calculating the variance across the $n$ samples of $\mathcal{D}_c^i$.
Then we compute the per-location uncertainty map by summing the variances across all classes:
\begin{equation}
\mathcal{U}(x, y) = \sum_{c \in C} \sigma^2_c(x, y).
\end{equation}
This formulation provides a spatially-resolved uncertainty estimate, where higher values of $\mathcal{U}$ indicate greater variability in class predictions across samples, highlighting regions of increased uncertainty in the local HD map.
The values of $\mathcal{U}$ are derived from variance measures and are not normalized; consequently, they should be interpreted as qualitative indicators.

\section{Experiments} \label{sec:experiments}

\subsection{Dataset and Evaluation Metrics}\label{sec:metrics}
We conduct our experiments on the nuScenes dataset \cite{nuscenes}, which provides data points at 2 Hz. 
Each data point includes images from six monocular cameras, $I$, and vectorized GT sets (\ie, $\mathcal{P}_{\mathrm{div}}$, $\mathcal{P}_{\mathrm{bound}}$, and $\mathcal{P}_{\mathrm{ped}}$ for lane dividers, lane boundaries, and pedestrian crossings, respectively).
We use the StreamMapNet training and validation split without geospatial overlaps \cite{streammapnet_2024}.
The graph decoding task is evaluated using Average Precision (AP) and mean Average Precision (mAP). 
For single-sample and $n$-sample rasterized predictions, we measure the True Positive Rate (TPR) and False Positive Rate (FPR) for various operational points and compare them using the Receiver Operating Characteristic (ROC) curve, and Area Under the Curve (AUC).
Inference time is measured in Frames Per Second (FPS).

\subsection{Experimental Setup}\label{sec:implementation}
We adopt the StreamMapNet training configuration with 24 epochs and a batch size of 1.
The model is trained in parallel on 8 NVIDIA V100 GPUs.
For optimization, we employ the AdamW optimizer with a cosine annealing schedule and a learning rate of $2 \times 10^{-4}$.
The dimensions of the BEV grid are set to $100 \times 50$, covering a perception range of $\SI{60}{\meter} \times \SI{30}{\meter}$. 
For diffusion, we choose a cosine noise scheduler and set $T=1000$.
Inference uses $\eta=0.5$ and $k=5$ diffusion steps.
For experiments generating multiple samples, we set $n=10$.
The $\mathrm{Rasterize}$ operator uses pixel width 1 on polylines with prediction score $>0.4$.
The Gaussian filter uses $g = 5$ (\SI{3}{\meter}) and $\sigma=1$ (\SI{0.6}{\meter}).

\subsection{Baseline Models}\label{sec:baseline}
We use StreamMapNet \cite{streammapnet_2024} as reference architecture and primary baseline. 
Other baselines include PolyDiffuse \cite{Chen2023PolyDiffuse}, the only diffusion-based method for online vectorized map construction, SQD-MapNet \cite{wang2024stream}, which performs a similar strategy of denoising on queries, and common methods for vectorized map construction, including VectorMapNet \cite{vectormapnet_2023}, MapTR \cite{maptr_2023}, MapTRv2 \cite{maptrv2_2023}, MapVR \cite{zhang2024mapvr}, and MGMap \cite{liu2024mgmap}.

\subsection{Quantitative Results of Model}\label{sec:quantitative_results}
\Cref{tab:table_main_result_new_split} shows the qualitative results.
MapDiffusion reaches \SI{35.6}{\percent}~mAP, a \SI{5.3}{\percent} relative improvement over the StreamMapNet baseline with \SI{33.8}{\percent}~mAP.
Notably, this performance is achieved despite MapDiffusion operating without learned queries and lacking temporal aggregation in the decoder, underscoring its efficacy in generating high-quality map samples under these constraints.
The model remains highly efficient, achieving real-time performance at 8.0~FPS with five diffusion steps. In a single-step configuration, it matches StreamMapNet with 12.8~FPS.
MapDiffusion also outperforms common baselines including VectorMapNet \cite{vectormapnet_2023}, MapTR \cite{maptr_2023}, MapTRv2 \cite{maptrv2_2023}, MapVR \cite{zhang2024mapvr}, and MGMap \cite{liu2024mgmap}.
We run the public implementation of SQD-MapNet \cite{wang2024stream} on the new nuScenes split with batch size 1 and get \SI{33.1}{\percent}~mAP, which ranks it below our MapDiffusion approach.
PolyDiffuse \cite{Chen2023PolyDiffuse} did not release code to reproduce the results on the new nuScenes split, but they report a result below StreamMapNet on the original split. 
Given our \SI{5.3}{\percent} relative improvement over StreamMapNet, we assume from transitive reasoning that MapDiffusion outperforms PolyDiffuse. Very recent works reach beyond the performance of StreamMapNet (\eg, MapQR \cite{liu2024leveraging}, HIMap \cite{zhou2024himap}), but do not report results on the new nuScenes split. 
This work intends to show the general possibility of using diffusion for generative map construction and the benefit of deriving an uncertainty estimate.
Most importantly, our paradigm can be applied to these works as well.

{\begin{table}[tbp]
    \centering
    \vspace{0.6em}  
    \caption{Performance of MapDiffusion compared to various baselines at perception range $\SI{60}
    {\meter} \times \SI{30}{\meter}$ on nuScenes split without geospatial overlap \cite{streammapnet_2024}. $^*$ results from \cite{streammapnet_2024}, all other results are reproduced. AP thresholds are $\{0.5, 1.0, 1.5\}$.}
    \label{tab:table_main_result_new_split}
    \resizebox{\columnwidth}{!}{%
    \begin{tabular}{lccccc}
        Method & AP$_{\mathrm{ped}}$ &  AP$_{\mathrm{div}}$ & AP$_{\mathrm{bound}}$ & mAP \\ 
        \midrule
        VectorMapNet$^*$ \cite{vectormapnet_2023} & 15.8 & 17.0 & 21.2 & 18.0  \\
        MapTR \cite{maptr_2023}       & 7.5 & 23.0 & 35.8 & 22.1 \\
        MapVR \cite{zhang2024mapvr}       & 10.1 & 22.6 & 35.7 & 22.8 \\
        MGMap \cite{liu2024mgmap}       & 7.9 & 25.6 & 37.4 & 23.7 \\
        MapTRv2 \cite{maptrv2_2023}       & 16.2 & 28.7 & 44.8 & 29.9 \\
        SQD-MapNet \cite{wang2024stream}  & 31.6 & 27.4 & 40.4 & 33.1 \\
        StreamMapNet \cite{streammapnet_2024} &  31.2 & 27.3 & \textbf{42.9} & 33.8 \\
        MapDiffusion (ours) & \textbf{32.9} & \textbf{31.4} & 42.4 & \textbf{35.6} \\
    \end{tabular}%
    }
    \vspace{-4pt}
\end{table}
}

\subsection{Aggregating Samples}\label{sec:uncertainty_experiments}
All previous results were calculated from one sample. 
Experiments below show the benefit of multiple samples. 

\subsubsection{Refined Prediction}\label{sec:refined_prediction}
\begin{figure}
    \vspace{0.4em}  
    \includegraphics[width=0.975\columnwidth, left]{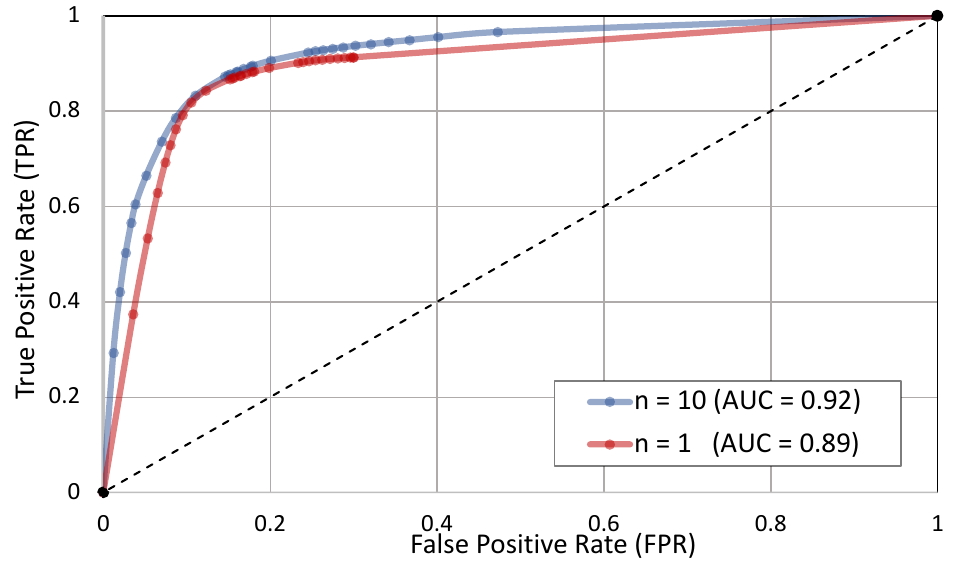}
    \caption{ROC curves for one sample (red) and 10 aggregated samples (blue). Dashed line is random classifier. AUC indicates that aggregated map is better than single-sample map.}
    \label{fig:roc_curve}
\end{figure}

Using the rasterized predicted class distribution $\mathcal{D}_c$ and the rasterized GT map ${\mathcal{M}}$, we compute the True Positive Rate and False Positive Rate for different binarization thresholds $b$.
The resulting ROC curves for $n=1$ and $n=10$ are visualized in \Cref{fig:roc_curve}.
The aggregated prediction from 10 samples is strictly better along the curve, confirming the hypothesized benefit of sampling multiple predictions from the distribution.
Accordingly, the AUC is 0.89 for $n=1$ and 0.92 for $n=10$, indicating a \SI{3.4}{\percent} relative improvement from aggregating multiple samples.
Please note that while this improvement is notable, we perform this evaluation primarily to show the information gain by generating multiple samples from the full distribution.
The true value lies in considering all plausible map configurations in the downstream planning module for more robust and uncertainty-aware decision-making.

\subsubsection{Uncertainty and Visibility}\label{sec:uncertainty_visibility}
\begin{figure}
    \centering
    \includegraphics[width=0.975\columnwidth, right]{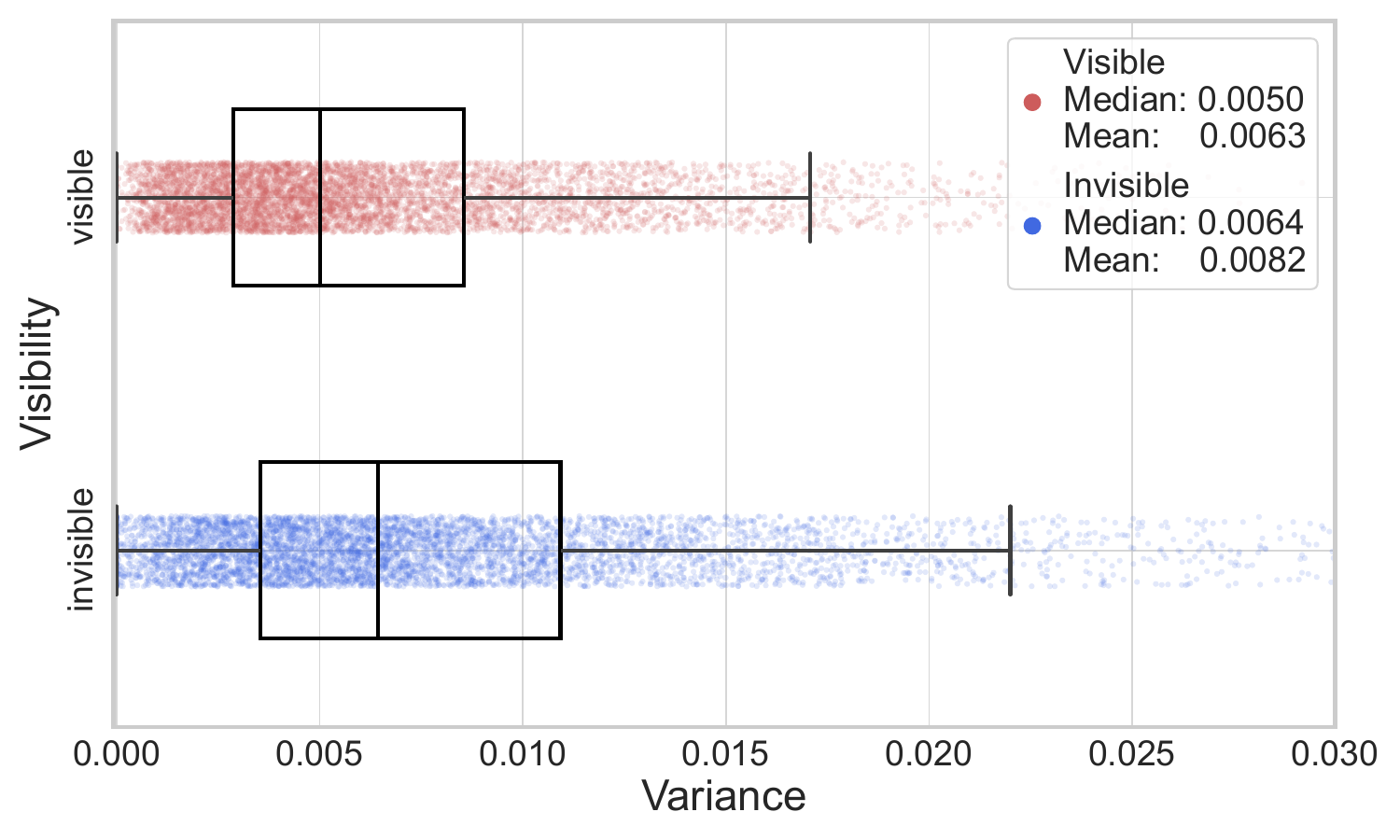}
    \caption{Relation between visibility and variance across predicted map samples, which we use as an uncertainty estimate.}
    \label{fig:boxplot}
\end{figure}

We aim to show that the uncertainty maps $\mathcal{U}$ from our approach correspond to ambiguity in the real world. 
In that case, $\mathcal{U}$ is expected to estimate high uncertainty for areas that are not visible, for example, due to a vehicle occluding the camera's field of view.
We compute the spatial relation between our estimated uncertainty maps and GT visibility masks.
Since nuScenes does not provide visibility masks directly, we generate them from the Occ3D dataset \cite{tian2023occ3d}, which provides occupancy maps for the nuScenes dataset.
We exclude the ground layer and occupancy of type \enquote{flat} and project the 3D voxels into 2D BEV based on whether any of the voxels in the z direction are occluded.
Finally, we perform ray-tracing on the 2D projection to calculate our visibility masks.
For our evaluation, we calculate the mean uncertainty per traffic scene separately for visible areas and invisible areas and compare them.
We consider pixels that are visible (based on the visibility map) or part of the drivable road surface (based on the dense raster GT).
The distribution of the variances, which we use as uncertainty estimates, is shown in \Cref{fig:boxplot}.
The mean uncertainty for the visible area is 0.0063.
For the invisible area, it is 0.0082, which is \SI{31}{\percent} higher.
We conduct a one-sided t-test and find that uncertainty in invisible areas is significantly higher than in visible areas ($\alpha = 0.01$, $p < 0.001$).
It is important to note that the MapDiffusion model performs temporal aggregation in BEV space and hence has access to more spatial features than currently visible, reducing the uncertainty in invisible areas that were visible in previous time steps.
Therefore, the relation between uncertainty and visibility is assumed to be even higher for a model with no temporal aggregation.

\subsection{Qualitative Results}\label{sec:qualitative_results}
\begin{figure*}
    \centering
    \vspace{0.4em}  
    \begin{subfigure}[t]{0.496\textwidth}
        \centering
        \includegraphics[trim=0cm 0cm 0cm 0cm, clip, width=\textwidth]{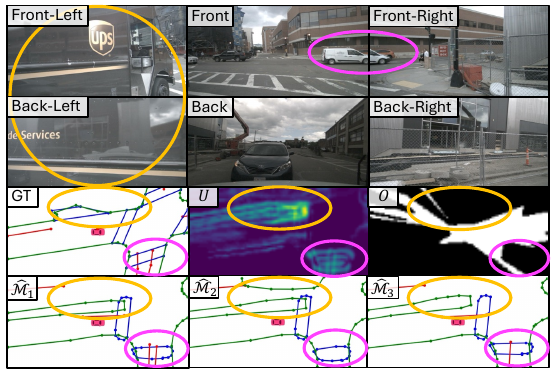}
        \caption{Occlusion from the delivery truck on the ego's left is shown with orange circles and results in different $\mathrm{bound}$ predictions. Occlusion from the white van on the ego's front-right is shown with pink circles and results in varying $\mathrm{div}$ and $\mathrm{ped}$ predictions.}
        \label{fig:qualitative_result_1}
    \end{subfigure}
    \hfill
    \begin{subfigure}[t]{0.496\textwidth}
        \centering
        \includegraphics[trim=0cm 0cm 0cm 0cm, clip, width=\textwidth]{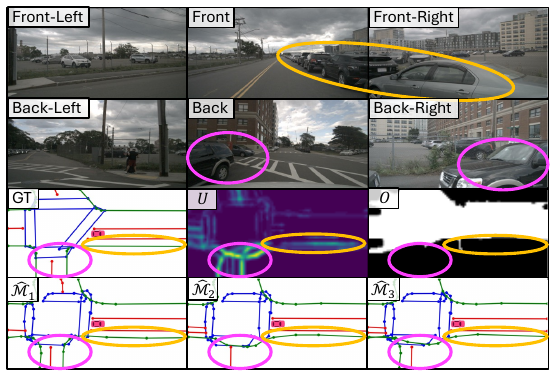}
        \caption{Occlusion from parked cars on the ego's right is shown with orange circles and results in different $\mathrm{bound}$ predictions.
        The occluded intersection at the ego's back-right is shown with pink circles and results in varying $\mathrm{bound}$ and $\mathrm{ped}$ predictions.}
        \label{fig:qualitative_result_2}
    \end{subfigure}
    \caption{Two qualitative results of MapDiffusion. The top two rows show the 6 camera views. The third row shows the GT (left), uncertainty $\mathcal{U}$ (center), and occlusion (right) maps. The bottom rows show 3 predicted samples $\hat{\mathcal{M}}_{1:3}$, where green is $\mathrm{bound}$, red is $\mathrm{div}$, and blue is $\mathrm{ped}$.}
    \label{fig:qualitative_result}
\end{figure*}

\Cref{fig:qualitative_result} shows qualitative results for two traffic scenes.
Multiple map construction samples are visualized for each traffic scene to illustrate the sampling variance.
The predicted map demonstrates high accuracy in visible areas, validating its state-of-the-art quantitative performance.
In both scenes, the variance across sampled map predictions is high in occluded areas, showcasing the relation between uncertainty and perceptual ambiguity.

\subsection{Ablation Studies}\label{sec:Ablation Studies}
We perform ablation studies to examine the efficacy of design decisions, including choice of diffusion parameters, pretraining of the BEV encoder, and padding strategies.

\subsubsection{Diffusion Parameters}
{\begin{table}
    \centering
    \vspace{0.6em}  
    \caption{Ablation on Diffusion Parameters $k$, $\eta$, $\tau$.}
    \label{tab:diffusion_parameters}
    \resizebox{\columnwidth}{!}{%
    \begin{tabular}{cllcccccc}
        Steps $k$ & $\eta$ & $\tau$ & FPS & AP$_{\mathrm{ped}}$ & AP$_{\mathrm{div}}$ & AP$_{\mathrm{bound}}$ & mAP \\
        \midrule
        1 & 0.5 & 0.5 & 12.8 & 32.3 & 30.8 & 42.5 & 35.2 \\
        2 & 0.5 & 0.5 & 11.1 & 32.3 & 31.4 & 42.5 & 35.4 \\
        3 & 0.5 & 0.5 & 9.9  & 32.9 & 31.2 & 42.5 & 35.5 \\
        4 & 0.5 & 0.5 & 8.8  & 32.9 & 31.5 & 42.5 & \textbf{35.6} \\
        5 & 0.5 & 0.5 & 8.0  & 32.9 & 31.4 & 42.4 & \textbf{35.6} \\
        \midrule
        5 & 0.1  & 0.5 & 8.0  & 32.7 & 31.3 & 42.3 & 35.4 \\
        5 & 0.3  & 0.5 & 8.0  & 32.6 & 31.2 & 42.3 & 35.4 \\
        5 & 0.5  & 0.5  & 8.0  & 32.9 & 31.4 & 42.4 & \textbf{35.6} \\
        5 & 0.7  & 0.5 & 8.0  & 32.6 & 31.4 & 42.4 & 35.5 \\
        5 & 0.9  & 0.5 & 8.0  & 32.5 & 31.3 & 42.4 & 35.4 \\
        \midrule
        5 & 0.5  & 0.1 & 8.0  & 30.9 & 27.8 & 39.3 & 32.7 \\
        5 & 0.5  & 0.3 & 8.0  & 32.5 & 31.2 & 42.3 & 35.3 \\
        5 & 0.5  & 0.5  & 8.0  & 32.9 & 31.4 & 42.4 & \textbf{35.6} \\
        5 & 0.5  & 0.7 & 8.0  & 33.0 & 31.1 & 42.3 & 35.5 \\
        5 & 0.5  & 0.9 & 8.0  & 32.7 & 30.6 & 42.1 & 35.1 \\
    \end{tabular}%
    }
\end{table}}
{\begin{table}
    \centering
    \caption{Ablation on query padding strategies. Models were trained for 12 epochs with a pretrained, frozen BEV encoder.}
    \label{tab:table_padding_ablation}
    \resizebox{0.8\columnwidth}{!}{%
        \begin{tabular}{lcccc}
        Padding & AP$_{\mathrm{ped}}$ & AP$_{\mathrm{div}}$ & AP$_{\mathrm{bound}}$ & mAP \\
        \midrule
        Repeat & 23.6 & 22.8 & 35.8 & 27.4 \\
        Zero & 25.2 & 28.3 & 37.8 & 30.5 \\
        Smooth &  24.8 & 27.8 & 38.1 & 30.2 \\
        Gaussian & 26.3 & 27.8 & 38.6 & \textbf{30.9} \\
        Uniform & 25.0 & 27.7 & 37.3 & 30.0 \\
        \end{tabular}%
    }
\end{table}}

We evaluate the number of diffusion steps, the $\eta$ parameter in DDIM sampling \cite{song2020denoising}, and the query threshold $\tau$.
The results are shown in \Cref{tab:diffusion_parameters}.
For the number of steps, we see better performance on the map construction task with more diffusion steps, saturating at around 5.
The number of diffusion steps has an effect on the runtime since it requires sequential execution of the denoising module, a typical downside of diffusion models.
Our approach addresses this issue by excluding the learned BEV encoder from the diffusion process, so the latent BEV grid only has to be computed once.
On an NVIDIA A10 GPU, performing an additional denoising step adds around \SI{12}{\milli\second}. This is a \SI{+15}{\percent} increase from the baseline inference time that has just one decoder pass.
The five diffusion steps necessary to achieve saturated performance increase the total time by \SI{60}{\percent}, achieving 8.0~FPS.
It is important that only the number of diffusion steps $k$ increases the latency. 
Sampling the distribution with $n$ predictions can be done in parallel.

The $\eta$ parameter has only a minor influence, and we get the best results for $\eta = 0.5$.
The query threshold $\tau$, which determines which queries are kept for the next diffusion step, has a stronger impact.
Keeping almost all queries ($\tau = 0.1$) has the worst performance with \SI{32.7}{\percent}~mAP.
The best performance reaches \SI{35.6}{\percent}~mAP for $\tau = 0.5$.
Dropping most queries ($\tau = 0.9$) degrades performance again to \SI{35.1}{\percent}~mAP.
Overall, given an mAP of around \SI{35}{\percent} for most settings, we find that MapDiffusion is robust to hyperparameter choices.
While even one diffusion step already reaches a good result, more steps are expected to increase sample variance, which is beneficial for capturing the full distribution and also generating $\mathcal{U}$.
Based on the ablation results, we choose the number of diffusion steps to be 5, $\eta=0.5$, and $\tau=0.5$.

\subsubsection{Pretraining of BEV Encoder}\label{sec:pretraining}
We assess the benefit of pretraining the BEV encoder. 
First, we train the full MapDiffusion model with randomly initialized weights, reaching \SI{35.6}{\percent}~mAP. 
We then train a new MapDiffusion model with the frozen pre-trained BEV encoder.
While the optimization is faster with a pre-trained BEV encoder (\SI{25.7}{\percent}~mAP vs. \SI{17.5}{\percent}~mAP after 6 epochs, and convergence around 12 epochs), the resulting model reaches only \SI{31.7}{\percent}~mAP.
Hence, we opt for training from scratch for all final experiments. 
We use the pretraining method exclusively for ablations, such as ablating the padding strategy, and train for 12 epochs there for efficiency reasons.

\subsubsection{Padding Strategy}\label{sec:padding_ablation}
We compare the following strategies for query padding.
\enquote{Repeat}: repeats existing polylines, \enquote{Zero}: zero values for all additional polylines, \enquote{Smooth}: smooth random polylines (both straight and curved) or polygons (\eg, elliptical shapes), \enquote{Gaussian}: Gaussian noise with $\mu=0.5$ and $\sigma=0.25$ clipped to [0,1] (range of normalized GT), and \enquote{Uniform}: uniform noise with the boundaries $[0,1]$.
For efficiency, we train them with a pre-trained BEV encoder for 12 epochs (see \Cref{sec:pretraining}).
The results are shown in \Cref{tab:table_padding_ablation}.
\enquote{Gaussian} performs best.
\enquote{Zero}, \enquote{Smooth}, and \enquote{Uniform} perform reasonably well.
\enquote{Repeat} surprisingly performs much worse, likely due to the model confusing the multiple accurate polylines.

\section{Conclusion} \label{sec:conclusion}

MapDiffusion is a novel approach that leverages generative diffusion for online vectorized HD map construction in autonomous driving. 
By integrating a diffusion-based denoising decoder with a learned BEV encoder, MapDiffusion predicts multiple plausible map representations from noisy initial queries.
This sampling of the map distribution can also provide spatial uncertainty estimates. 
Experiments on the nuScenes dataset demonstrated that MapDiffusion achieves state-of-the-art performance, with a relative improvement of \SI{5}{\percent} over the StreamMapNet baseline, even without access to queries that are learned or temporally aggregated from the previous frame. 
Additionally, by sampling outputs, our approach enhances prediction accuracy and generates useful uncertainty maps.
Ablation studies revealed optimal configurations for diffusion parameters, query padding strategies, and the impact of pretraining the BEV encoder. 
Moreover, we showed that the uncertainty maps generated by MapDiffusion estimate significantly higher uncertainty in invisible areas, highlighting their practical relevance for real-world applications. 
In conclusion, MapDiffusion establishes the new state of the art for online map construction on nuScenes and emphasizes the potential of generative diffusion models in online mapping tasks, proving their ability to enhance accuracy, reliability, and robustness. 
This generic framework can be applied to other models, improving their performance while providing valuable uncertainty estimates, paving the way for safer and more robust autonomous driving systems.

\pagebreak

\bibliography{Literature}{}
\bibliographystyle{IEEEtran}

\end{document}